%% file: Block_HRI_2021.tex
\documentclass[sigconf]{acmart}

\AtBeginDocument{%
  \providecommand\BibTeX{{%
    \normalfont B\kern-0.5em{\scshape i\kern-0.25em b}\kern-0.8em\TeX}}}

\usepackage{graphicx}
\usepackage{amsmath}
\usepackage{tabularx}
\usepackage{booktabs}
\usepackage{scrtime}
\usepackage[dont-mess-around]{fnpct}
\usepackage{etoolbox}
\usepackage{balance}
\copyrightyear{2021} 
\acmYear{2021} 
\setcopyright{acmlicensed}
\acmDOI{10.1145/3434073.3444656}

\acmConference[HRI '21] {Proceedings of the 2021 ACM/IEEE International Conference on Human-Robot Interaction}{March 8--11, 2021}{Boulder, CO, USA}
\acmBooktitle{Proceedings of the 2021 ACM/IEEE International Conference on Human-Robot Interaction (HRI '21), March 8--11, 2021, Boulder, CO, USA}
\acmPrice{15.00}
\acmISBN{978-1-4503-8289-2/21/03}

\acmSubmissionID{hrifp1096}

\begin{document}
\fancyhead{}

\title{The Six Hug Commandments: Design and Evaluation of a Human-Sized Hugging Robot with Visual and Haptic Perception}
\renewcommand{\shorttitle}{The Six Hug Commandments}

\author{Alexis E. Block}
\orcid{0000-0001-9841-0769}
\affiliation{
  \institution{MPI-IS and ETH Z\"{u}rich}
  \city{Stuttgart}
  \state{Germany}
}

\author{Sammy Christen}
\orcid{0000-0002-3511-8565}
\affiliation{%
  \institution{ETH Z\"{u}rich}
  \city{Z\"{u}rich}
  \country{Switzerland}}

\author{Roger Gassert}
\orcid{0000-0002-6373-8518}
\affiliation{%
 \institution{ETH Z\"{u}rich}
 \city{Z\"{u}rich}
 \country{Switzerland}}

\author{Otmar Hilliges}
\orcid{0000-0002-5068-3474}
\affiliation{%
  \institution{ETH Z\"{u}rich}
  \city{Z\"{u}rich}
  \country{Switzerland}
}

\author{Katherine J. Kuchenbecker}
\orcid{0000-0002-5004-0313}
\affiliation{%
  \institution{MPI for Intelligent Systems}
  \city{Stuttgart}
  \country{Germany}}

\renewcommand{\shortauthors}{Block et al.}
\newcommand{\vecb}[1]{\textbf{#1}}
\newcommand{\vect}[1]{{#1}}
\newcommand{\note}[3]{{\color{#2}[#1: #3]}}
\newcommand{\SC}[1]{\note{SC}{blue}{#1}}
\input{./sections/00_abstract}




\maketitle

\input{./sections/01_introduction}
\input{./sections/02_related_work}
\input{./sections/03_hypotheses}
\input{./sections/04_system}
\input{./sections/05_online_study}
\input{./sections/06_in_person_study}
\input{./sections/07_discussion}
\input{./sections/08_conclusion}

\begin{acks}
This work is partially supported by the Max Planck ETH Center for Learning Systems and the IEEE Technical Committee on Haptics. The authors thank Bernard Javot, Michaela Wieland, Hasti Seifi, Felix Gr\"{u}ninger, Joey Burns, Ilona Jacobi, Ravali Gourishetti, Ben Richardson, Meike Pech, Nati Egana, Mayumi Mohan, and Kinova Robotics for supporting various aspects of this research project. 
\end{acks}

\clearpage
\bibliographystyle{ACM-Reference-Format}
\balance
\bibliography{references}

\input{./sections/appendix}

\end{document}

%% file: sections/00_abstract.tex
\begin{abstract}  

Receiving a hug is one of the best ways to feel socially supported, and the lack of social touch can have severe negative effects on an individual's well-being. Based on previous research both within and outside of HRI, we propose six tenets (``commandments'') of natural and enjoyable robotic hugging: a hugging robot should be soft, be warm, be human sized, visually perceive its user, adjust its embrace to the user's size and position, and reliably release when the user wants to end the hug. Prior work validated the first two tenets, and the final four are new. We followed all six tenets to create a new robotic platform, HuggieBot~2.0, that has a soft, warm, inflated body (HuggieChest) and uses visual and haptic sensing to deliver closed-loop hugging. We first verified the outward appeal of this platform in comparison to the previous PR2-based HuggieBot 1.0 via an online video-watching study involving 117 users. We then conducted an in-person experiment in which 32 users each exchanged eight hugs with HuggieBot~2.0, experiencing all combinations of visual hug initiation, haptic sizing, and haptic releasing. The results show that adding haptic reactivity definitively improves user perception a hugging robot, largely verifying our four new tenets and illuminating several interesting opportunities for further improvement. \looseness-1

\end{abstract}

%% file: sections/01_introduction.tex
\section{Introduction}
\label{Introduction}

\begin{figure}[t]
\includegraphics[width=.7\columnwidth, trim = {0.35cm 0.35cm 0.35cm 50.5cm}, clip]{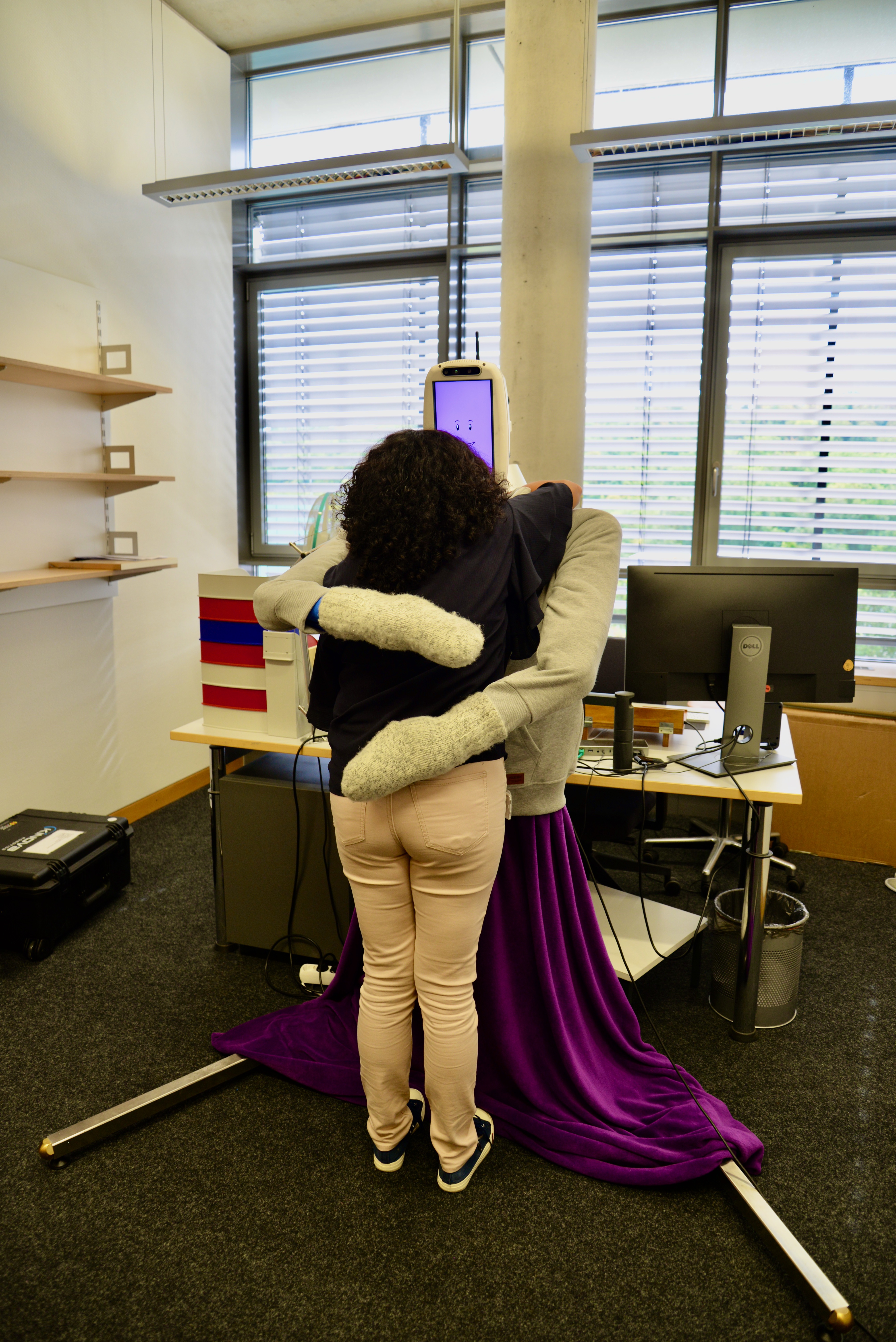}
\vspace{-0.25cm}
\caption{A user hugging HuggieBot 2.0.}
\label{fig:Hug}
\vspace{-0.65cm}
\end{figure}

Hugging has significant social and physical health benefits for humans. Not only does a hug help lower blood pressure, alleviate stress and anxiety, and increase the body's levels of oxytocin, but it also provides social support, increases trust, and fosters a sense of community and belonging \cite{HuggingHealthBenefits}. Social touch in a broader sense is also vital for maintaining many kinds of relationships among humans and primates alike \cite{TopographyOfSocialTouch}; hugs seem to be a basic evolutionary need.  They are therefore highly popular! An online study conducted in 2020 polled 1,204,986 people to find out ``what is the best thing?'' Hugs earned fifth place out of 8,850 things, behind only sleep, electricity, the Earth's magnetic field, and gravity \cite{bestthing}. The absence of social touch can have detrimental effects on child development \cite{SocialTouchDevelopment}. Unfortunately, ever more interactions are happening remotely and online, 
especially during this unprecedented time of physical distancing due to COVID-19. An increasing number of people are suffering from loneliness and depression due to increased workload and population aging \cite{DepressedTeensFromSocialMedia,InternetAndDepression}. Our long-term research goal is to determine the extent to which we can close the gap between the virtual and physical worlds via \textit{hugging robots that provide high-quality social touch}. \looseness-1

Making a good hugging robot is difficult because it must understand the user's nonverbal cues, realistically replicate a human hug, and ensure user safety. We believe that such robots need multi-modal perception to satisfy all three of these goals, a target no previous system has reached. Some approaches focus primarily on safety, providing the user with the sensation of being hugged without being able to actively reciprocate the hugging motion \cite{tsetserukou2010haptihug,teh2008huggypajama,duvall2016active}. Conversely, other researchers focus on providing the user with an item to hug, but that item cannot hug the user back \cite{HuggablePillowWithPhone,TheHuggable,disalvo2003hug}. Other robotic solutions safely replicate a hug, but they are teleoperated, meaning they have no perception of their user and require an additional person to control the robot any time a user wants a hug \cite{hedayati2019hugbot,shiomi2017hug,Disney}. Finally, some robots have basic levels of perception but are not fully autonomous or comfortable \cite{block2019softness,miyashita2004human}. 
Section~\ref{RelatedWork} further details prior research in this domain.\looseness-1


To tackle the aforementioned goal of safely delivering pleasant hugs, we propose the \textit{six tenets  (``commandments'') of robotic hugging}: a hugging robot should be soft, warm, and sized similar to an adult human, and it should see and react to an approaching user, adjust automatically to that user's size and position while hugging, and reliably respond when the user releases the embrace. After presenting these tenets and our accompanying hypotheses in Section~\ref{Hypotheses}, we use the tenets to inform the creation of HuggieBot 2.0, a novel humanoid robot for close social-physical interaction, as seen in Fig.~\ref{fig:Hug} and described in Section~\ref{SystemDesign}. HuggieBot 2.0 uses computer vision to detect an approaching user and automatically initiate a hug based on their distance to the robot.  It also uniquely models hugging after robot grasping, using two slender padded arms, an inflated torso, and haptic sensing 
to automatically adjust to the user's body and detect user hug initiation and termination. HuggieBot 2.0 is the \textit{first human-sized hugging robot with visual and haptic perception for closed-loop hugging}. \looseness-1

We then seek to validate the four new tenets by conducting two experiments with HuggieBot 2.0. First, we confirmed user preference for the created platform's physical size, visual appearance, and movements through a comparative online study, as described in Section~\ref{Online User Study}.  We then conducted an in-person study (Section~\ref{In Person User Study}) to understand how HuggieBot 2.0 and its three new perceptual capabilities (vision, sizing, and release detection) affect user opinions. Section~\ref{Discussion} discusses the study results, which show that the six tenets significantly improve user perception of hugging robots. Section~\ref{Conclusion} discusses the limitations of our approach and concludes the paper.



%% file: sections/02_related_work.tex
\section{Related Work}
\label{RelatedWork}

\subsubsection*{\textbf{Using Vision for Person Detection}}
One challenge of accurate and safe robotic hugging is detecting a user's desire for a hug. Many researchers solve this problem by using a remote operator to activate the hug \cite{disalvo2003hug,Disney,Hugvie,HuggablePillowWithPhone}. This form of telehug is not a universal approach because it requires a hugging partner to be available at the exact moment a user would like the comfort of a hug. Having the user press a button is a simpler alternative but differs greatly from human-human hugging. One method that could allow robots to provide hugs autonomously is to detect an approaching user via computer vision. Human detection has long been of interest in many research fields, including autonomous driving \cite{yurtsever2019}, surveillance and security \cite{manoranjan2013surveillance}, computer vision \cite{viola01cv}, and human-robot interaction \cite{vo2014hri}. Early works focus mostly on finding a representative feature set that distinguishes the humans in the scene from other objects. Different methods for the feature extraction have been proposed, such as using Haar wavelets \cite{wavelet1997}, histograms of oriented gradients (HOG) \cite{dalal2006}, and covariance matrices \cite{tuzel2007}. Several approaches try to combine multiple cues for person detection; for example, Choi et al. \cite{choi2011} and Vo et al. \cite{vo2014hri} combine the Viola-Jones face detector \cite{violajones2001} and an upper-body detector based on HOG features \cite{dalal2006}. In computer vision, deep-learning-based systems relying on convolutional neural networks (CNNs) are often used for person detection. However, the computational cost of these detection pipelines is very high, so they are not suited for use on a real-time human-robot interaction platform. The recent decrease of the computational cost of improved models, e.g., \cite{Liu_2016}, has facilitated their adaptation to robot platforms. Hence, we integrate such a model into our pipeline so that HuggieBot 2.0 can recognize an approaching user to initiate a hug with minimal on-board computational load.\looseness-1

\subsubsection*{\textbf{Hugging as a Form of Grasping}}
Once the user arrives, safely delivering a hug is challenging for robots because users come in widely varying shapes and sizes and have different preferences for hug duration and tightness. 
No existing hugging robots are equipped to hug people adaptively. We propose looking to the robotics research community to find a solution. Grasping objects of varying shape, size, and mechanical properties is a common and well-studied problem, e.g., \cite{Romano11-TRO-Grasp,SpiersGrasping,TwoFingeredGrasp,GripSelection}. Therefore, we look at hugging as a large-scale two-fingered grasping problem, where the item to be grasped is a human body. For example, the BarrettHand, a commercially available three-fingered gripper, automatically adjusts to securely grasp widely varying objects by closing all finger joints simultaneously and then stopping each joint individually when that joint's torque exceeds a threshold \cite{BarrettHand}.\looseness-1  
The robot arms used for HuggieBot 2.0 have torque sensors at every joint, making this torque-thresholded grasping method an ideal way to achieve a secure embrace that neither leaves air gaps nor applies excessive pressure to the user's body. Torque sensors can also enable the robot to feel when the user wishes to leave the embrace.\looseness-1


\subsubsection*{\textbf{Previous Hugging Robots}}
In recent years, there has been a dramatic increase in the number of researchers developing and studying hugging robots. High press coverage shows this topic is of great interest to the general public. Interestingly, researchers are taking many different approaches to create robotic systems that can receive and/or give hugs. Smaller hugging robots have been created to provide comfort, but they typically cannot actively hug the user back. The Huggable is a small teddy bear to accompany children during long stays in the hospital \cite{TheHuggable}. The Hug is a pillow whose shape mimics a child wrapping his or her limbs around an adult \cite{disalvo2003hug}. Hugvie is also a pillow users can hug; a cellphone inside lets the user talk to a partner while he or she hugs the pillow \cite{HuggablePillowWithPhone,Hugvie}. Their small size makes these huggable systems inherently safer than larger devices, but it also prevents them from providing the benefits of social touch to the user because they cannot administer deep touch pressure therapy \cite{Edelson1999}.\looseness-1

Teleoperated hugging robots have also been created, and they can be closer to human size. Some research groups focus on non-anthropomorphic solutions like using a large panda or teddy bear stuffed animal to hide the mechanical components \cite{hedayati2019hugbot,shiomi2017hug,shiomi2017robot}. These robots all require that either an operator or partner is available at the time any user wants a hug. They also may not be very comfortable for the user because the robots are unable to stand on their own; the user must crouch or crawl to get a hug from the robot. Disney also patented a teleoperated hugging robot to be similar to its famous character Baymax \cite{Disney}, though a physical version has not been reported. Negatively, none of these robots seem to have the ability to adjust their embrace to the size or location of the user. In addition to creating an appropriately sized hugging robot, all of these researchers covered the robot's rigid components with soft materials to create an enjoyable contact experience for their users. In contrast, \citet{miyashita2004human} previously created a robot that measures the distance between the robot and user with range sensors to initiate the hug sequence; this robot has a hard surface that may be uncomfortable to touch, and its inverted pendulum design appears to use the human to balance during the hug.\looseness-1

Block and Kuchenbecker added padding, heating pads, and a soft tactile sensor to a Willow Garage Personal Robot 2 (PR2) to enable it to exchange hugs with human users \cite{block2019softness,block2018emotionally}. The experimenter manually adjusted the robot to match the height and size of each user, a process that takes time and prevents spontaneous hugging. She also initiated every hug for the user. The PR2 was successful in adapting to the user's desired hug duration through the use of the tactile sensor, but the user had to place his or her hand in a specific location on the PR2's back, which was not natural for all users, and the user had to press this sensor to tell the robot to release them. Some users also criticized the size and shape of this robot as being awkward to hug. On the positive side, this study showed that both softness and warmth are important for a robot to deliver good hugging experiences; we therefore incorporate these already validated elements as our first two tenets. From \citet{trovato2016hugging} we learned that softness alone is not enough for a hugging robot; people are more receptive to hugging a robot that is wearing clothing, so we placed suitable clothes on HuggieBot 2.0.\looseness-1

%% file: sections/03_hypotheses.tex
\section{Hugging Tenets and Hypotheses}
\label{Hypotheses}

We propose six tenets to guide the creation of future hugging robots. The first two were validated by Block and Kuchenbecker \cite{block2019softness}, and the other four are proposed and validated in this paper. We believe that a hugging robot should: 
\begin{itemize}
\item[T1.] be soft,
\item[T2.] be warm,
\item[T3.] be sized similar to an adult human,
\item[T4.] visually perceive and react to an approaching user, rather than requiring a triggering action such as a button press by the user, an experimenter, or a remote hugging partner,
\item[T5.] autonomously adapt its embrace to the size and position of the user's body, rather than hug in a constant manner, and 
\item[T6.] reliably detect and react to a user's desire to be released from a hug regardless of his or her arm positions. 
\end{itemize}

Building off the previously described research, this project seeks to evaluate the extent to which the six tenets benefit user perception of hugging robots. Specifically, we aim to test the following three hypotheses:
\begin{itemize}
\item[H1.] When viewing from a distance, potential users will prefer the embodiment and movement of a hugging robot that incorporates our four new tenets over a state-of-the-art hugging robot that violates the tenets. 
\item[H2.] Obeying the four new tenets during physical user interactions will significantly increase the perceived safety, naturalness, enjoyability, intelligence, and friendliness of a hugging robot. 
\item[H3.] Repeated hugs with a robot that follows all six tenets will improve user opinions about robots in general. \end{itemize}

%% file: sections/04_system.tex
\section{System Design and Engineering}
\label{SystemDesign}
We introduce a new human-sized hugging robot with visual and haptic perception. This platform is designed to have a friendlier and more comfortable appearance than previous state-of-the-art hugging robots. Building off feedback from users of prior robots (Section~\ref{RelatedWork}), we focused on six areas for the design of this new, self-contained robot: the frame, arms, inflated sensing torso, head and face, visual person detection, and software architecture.\looseness-1

\subsubsection*{\textbf{Frame}}
HuggieBot 2.0's core consists of a custom stainless steel frame with a v-shaped base. The robot's height can be manually adjusted if needed, and the shape of the base allows users to come very close to the robot, as seen in Fig.~\ref{fig:Hug}. The user does not need to lean over a large base to receive a hug, unlike the PR2-based hugging robot \cite{block2019softness}. The v-shaped base also increases the safety and stability of the robot by acting to counteract any leaning force imparted by a user. This large base and counterweight ensure that even a large user approaching at a high speed intending to make an incorrect impact with the robot will not flip it over, inflict injury upon themselves, or cause damage to the robot.\looseness-1
\subsubsection*{\textbf{Arms}}
Two 6-DOF Kinova JACO arms are horizontally mounted to a custom stainless steel bracket attached to the top of the metal frame. To create a more approachable appearance, the grippers of the JACO arms were removed, and large padded mittens were placed over the wrist joints that terminate each arm. The arms are controlled by commanding target joint angles; movement is quiet, and the joints are not easily backdrivable when powered. The torque sensors at each joint are continually monitored so that hugs can be automatically adjusted to each user's size and position. The second joint (shoulder pan) and third joint (elbow flex) on each arm stop closing individually when they surpass a torque threshold, which we empirically set to 10 Nm and 5 Nm, respectively. The joint torques are also used to detect when a user is pushing back against the arms with a torque higher than 20 Nm, indicating his/her desire to be released from a hug. To create a comfortable and enjoyable tactile experience for the user, we covered the arms in soft foam and a sweatshirt.\looseness-1
\subsubsection*{\textbf{HuggieChest: Inflatable Torso}}
\begin{figure}[t]
\includegraphics[width=0.7\columnwidth, trim = {0cm 15cm 0cm 15cm}, clip]{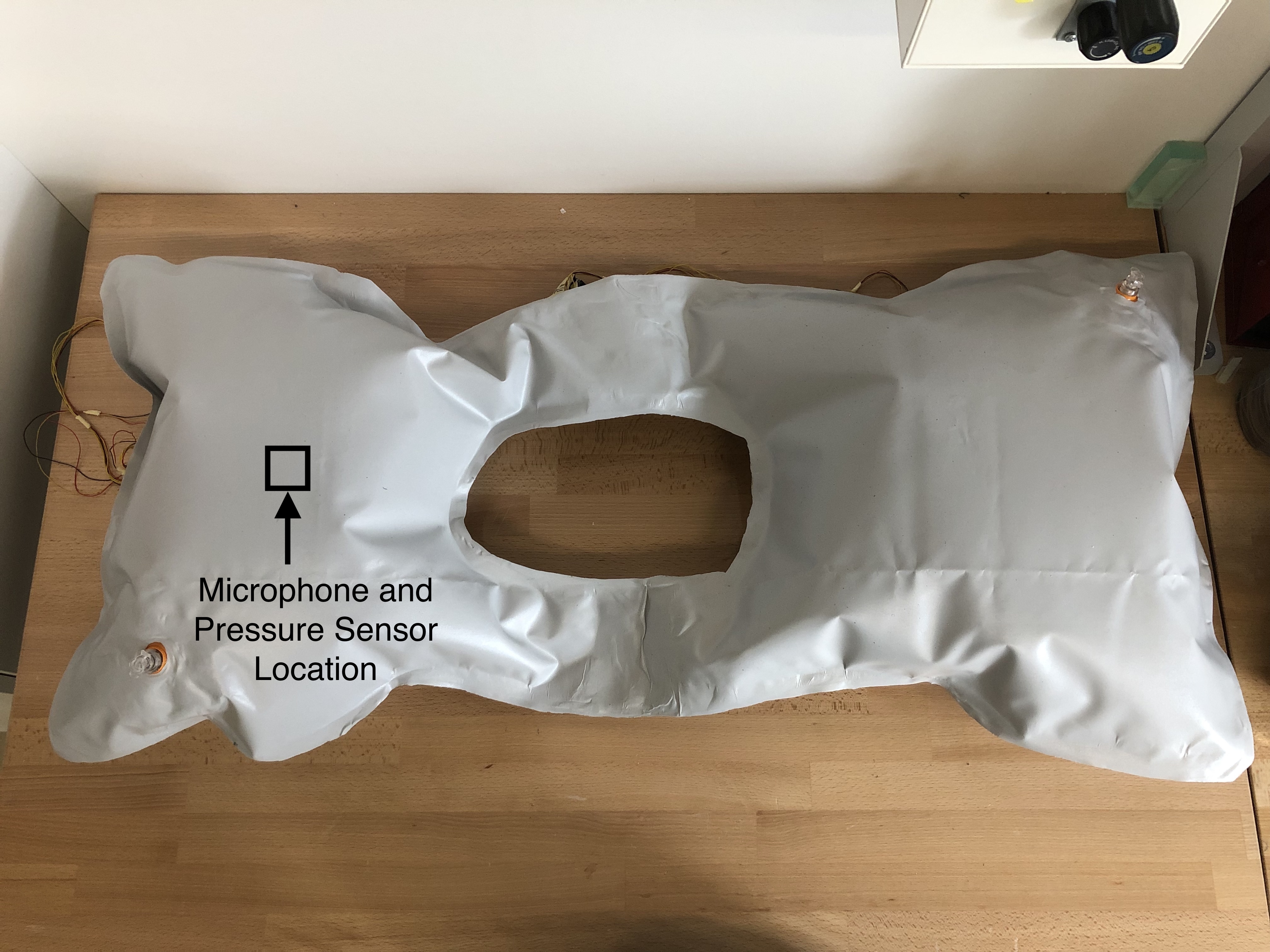}
\vspace{-0.3cm}
\caption{The inflated HuggieChest when lying flat. The two air chambers form the front and back of the robot's torso.}
\label{fig:HuggieChest}
\vspace{-0.3cm}
\end{figure}

We developed a simple and inherently soft inflatable haptic sensor to serve as the torso of our hugging robot, as pictured in Fig.~\ref{fig:HuggieChest}. The torso was constructed by both heat sealing and gluing (with HH-66 vinyl cement) two sheets of PVC vinyl to create an airtight seal. This chest has one chamber located in the front and another in the back. There is no airflow between the two chambers. Each chamber has a valve from an inflatable swim armband to inflate, seal, and deflate the chamber. Inside the chamber located on the back of the robot is an Adafruit BME680 barometric pressure sensor and an Adafruit electret microphone amplifier MAX4466 with adjustable gain. Both sensors were secured in the center of the chest on the inner wall of the chamber. The two sensors are connected to a single Arduino Uno micro-controller outside the chamber. The microphone and pressure sensor are sampled at 45 Hz, and the readings are sent over serial to the HuggieBot 2.0 computer for real-time processing. We originally tested with the same sensing capabilities in both chambers but did not find the information from the front chamber to be very useful; thus, no data are collected from the front chamber. This novel inflatable haptic sensor is called the HuggieChest.\looseness-1

The HuggieChest's shape was created by following a pattern of a padded vest that goes over the wearer's head and is secured with a belt around the waist. Because the HuggieChest is heat-sealed at the shoulders to stop airflow between the chambers and allow the chest to bend once inflated, the robot is unable to feel contacts in these locations. The HuggieChest is placed directly on top of the metal frame of HuggieBot 2.0. On top of the inflatable torso,  we put two Thermophore MaxHeat Deep-Heat Therapy heating pads (35.6 cm $\times$ 68.6 cm), which are attached together at one short edge with two shoulder straps. The sweatshirt is placed on top of the heating pads to create the final robot torso.\looseness-1
\subsubsection*{\textbf{Head and Face}}
We designed and 3D-printed a head to house a Dell OptiPlex 7050 minicomputer that controls the robot, a face screen, an RGB-D camera, the Arduino from the HuggieChest, a wireless JBL speaker, and cables. The head splits into two halves with a rectangular plate on each side that can be removed to access the inside. The final piece of the head is the front-facing frame, which secures the face screen and camera. The face screen is an LG LP101WH1 Display 10.1'' LCD screen with a 1366 $\times$ 768 resolution in portrait orientation. The screen displays faces based on designs created and validated for the Baxter robot \cite{fitter2016designing}.\looseness-1


\subsubsection*{\textbf{Vision and Person Detection}}
We use a commercially available Intel Realsense RGB-D camera with custom software to recognize an approaching person and initiate a hug. To this end, we integrate a deep-learning-based person detection module into our pipeline. The module consists of two parts. First, our software recognizes an approaching person using an open-source Robot Operating System (ROS) integration \cite{Odabasi2017} of Tensorflow's object detection library, which is based on the SSD mobilenet model \cite{Liu_2016}. In the next step, we utilize the camera's depth sensor to estimate the distance of the person to the robot. We use a sliding window to ensure a person is actively approaching the robot; we observe the distance measured from the depth sensor, which can be noisy, and check whether the mean distance decreases. This strategy prevents undesired hugs in case a person walks away from the robot. Once the person is actively approaching the robot, we initiate a hug as soon as a tuned distance threshold of 2.45~m is passed. This threshold 
was selected as it informs the robot the person is attempting to move from the social space into the robot's personal space \cite{Proxemics}. 
\subsubsection*{\textbf{Robot Software Architecture}}
The robot is controlled via ROS Kinetic. Each robot arm joint has both angle and torque sensors. A PID controller is used to control each joint angle over time.  The robot arms begin by moving to a home position. The camera module starts and waits for an approaching user. Upon detection, the robot asks the user, ``Can I have a hug, please?'' as in \cite{block2018emotionally}, while the robot's face changes to an opening and closing mouth. The specific hug it is supposed to run (with or without haptic sizing and release) is executed by commanding each joint to move at a fixed angular velocity toward a predetermined goal pose. For hugs without haptic sizing, the robot hugs in a one-size-fits-most manner, where the robot's second and third joints each close by 20$^\circ$. This pose was large enough such that it did not apply high forces to the bodies of any of our users; it was not adjusted for different subjects. For hugs with haptic sizing, the robot arms move toward a pose sized for a small user; we continually monitor each joint torque and stop a joint's movement if it exceeds the pre-set torque threshold. This method leads to automatic adjustment to the user's size (T5). 

The Arduino communicates the microphone and pressure sensor data from inside the back chamber of the HuggieChest to ROS over serial at 45~Hz. This data stream is analyzed in real time. The program first determines the ambient pressure and noise in the chamber by averaging the first 20 samples to create a baseline that accounts for different levels of inflation and noise. The chamber detects the user beginning to hug when the chamber's pressure increases by 50~kPa above the baseline pressure. Contact is determined to be broken, thus indicating that the user wants to be released, when the pressure returns to 10~kPa above the baseline pressure. The measured torques from the robot's shoulder pan and elbow flex joints are monitored continually during a hug. A haptic release is also triggered when any of these torques surpasses a threshold limit of 20~Nm. For a timed hug, rather than detecting the instant at which the user wants to be released, the robot waits 1 second after the arms fully close before releasing the user, so it is apparent to the user they are not in control of the duration of the hug. Overall, our proposed method of closed-loop hugging works on a higher level of abstraction than the low-level control, i.e., including both visual and haptic perception in the loop of the hugging process. The robot's haptic perception is two-fold: adjusting to the size of the user and sensing when he/she wants to be released. \looseness-1

%% file: sections/05_online_study.tex
\section{Online User Study}
\label{Online User Study}
We ran an online study to get feedback from a broad audience on the embodiment and movement of our robot as part of our user-centered design process, and to compare it to the PR2-based HuggieBot 1.0 \cite{block2019softness}. The main stimuli were two videos from \cite{block2019softness} along with two newly recorded videos of people hugging HuggieBot 2.0 with matched gender, enthusiasm, and timing; these videos are included as supplementary material for this paper. This study was approved by the Max Planck Society ethics council under the HI framework.\looseness-1 
\subsubsection*{\textbf{Participants}}
All participants for the online study were non-comp\-ensated English-speaking volunteers recruited via emails and social media. A total of 117 subjects took part in the online survey: 42.7\% male, 56.4\% female, and 0.9\% who identify as other. The participants ranged in age from 20 to 86 (M = 37.5, SD = 16.75). The majority of respondents indicated they had little (30.7\%) or no experience (43.6\%) interacting physically with robots.


\subsubsection*{\textbf{Procedures}}
After someone reached out to the experimenter and indicated interest in participating in the online study, the experimenter sent the subject an informed consent document by email. The participant read it thoroughly, asked any questions, and signed it and sent it back to the experimenter only if they wanted to participate. At this point, the user was assigned a subject number and sent a unique link to the online survey.

\begin{table}[t]
\caption{The questions participants answered after viewing or experiencing robot hugs.}
\vspace{-0.4cm}
\label{table:EachTrialSurvey}       
\begin{small}
\begin{tabularx}{\linewidth}{lX}
\hline\noalign{\smallskip}
This hug made the robot seem (unfriendly -- friendly)\\
This robot behavior seemed (unsafe -- safe)\\
This hug made the robot seem socially (stupid -- intelligent)\\
This hug interaction felt (awkward -- enjoyable)\\
This robot behavior seemed (unnatural -- natural)\\
\noalign{\smallskip}\hline
\end{tabularx}
\vspace{-.3cm}
\end{small}
\end{table} 

First, participants filled out their demographic information.
Next, they were shown two videos of an adult (one male and one female) hugging a robot labeled ``Robot A''.  Robot A was HuggieBot 2.0 for half of the participants and Block and Kuchenbecker's HuggieBot 1.0~\cite{block2019softness} for the other half.
Users could watch these videos as many times as they liked before answering several questions. Users described their first impressions of the robot they saw. Then, they answered the questions shown in Table \ref{table:EachTrialSurvey} on a 5-point Likert scale. Afterwards, there was an optional space for additional comments. Next, they were shown two videos of people hugging the other robot labeled ``Robot B'' under the same conditions as the first videos. 
Users answered the same questions for Robot B as they did for Robot A. Finally, since the videos were shot from behind the robot, users were shown frontal images of both robots posed in a similar manner. Participants were then asked ``In what ways is Robot A better than Robot B?'' and ``In what ways is Robot B better than Robot A?'' Then, they were asked to select which robot they would prefer to hug, Robot A or B, and why. 

After the first 40 participants, we noticed that several users were commenting on the purple fuzzy appearance of the PR2, rather than on the robots themselves. Therefore, we added a new section at the end of the survey, which was completed by the remaining 77 participants. This new section showed photographs of Robot A and Robot B plus two additional photographs showing HuggieBot 2.0 in different clothing.  In one of the new photographs, HuggieBot 2.0 wore a fuzzy purple robe similar to the fabric cover on the PR2, and in the other it wore its gray sweatshirt over this same fuzzy purple robe. Participants were asked which of these four robots they would prefer to hug and why.


\begin{figure}[t]
\includegraphics[trim = {2.5cm 2.5cm 1.3cm 4.5cm}, clip, width = \columnwidth]{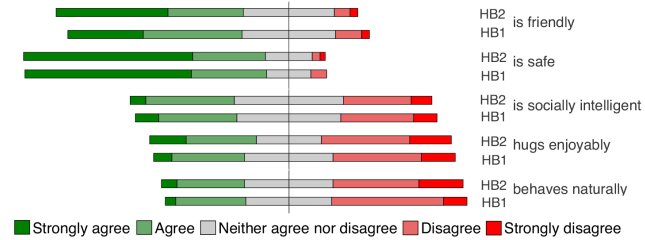}
\vspace{-0.8cm}
\caption{The responses to the five questions asked after users watched two videos of people hugging HuggieBot 2.0 (HB2) and HuggieBot 1.0 (HB1).}
\vspace{-0.4cm}
\label{fig:10questions}
\end{figure}

\subsubsection*{\textbf{Results}}
The responses to the five Likert-style questions from Table \ref{table:EachTrialSurvey} can be seen in Fig.~\ref{fig:10questions}. For all statistical analyses, we applied a Bonferroni alpha correction to $\alpha$ = 0.05 to determine significance and to account for the multiple comparisons. Because the data from these questions were non-parametric, we conducted a Wilcoxon signed-rank test. No statistically significant differences were found between the responses to any of these questions for the two robots. 

\begin{figure}[t]
\includegraphics[trim = {0.75cm 0.75cm 1.5cm 2.25cm}, clip, width = \columnwidth]{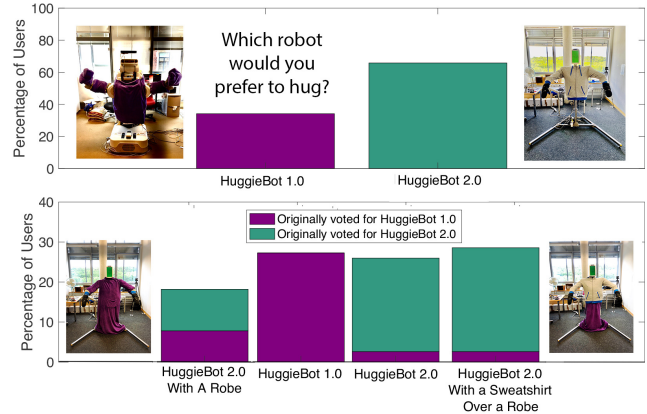}
\vspace{-0.8cm}
\caption{The breakdown of preferences when users had two choices (top) and four choices (bottom), with the associated images. The colors of the second plot show which robot the user preferred in the first selection round.}
\vspace{-0.4cm}
\label{fig:2VotingRounds}
\end{figure}

The responses to the first and second rounds of voting for which robot users would prefer to hug can be found in Fig.~\ref{fig:2VotingRounds}. 
We ran a Wilcoxon signed-rank test for the first round of voting and determined users would significantly prefer to hug our new robot over HuggieBot 1.0 (p < 0.001). In the second voting round, no significant preference was found between any of the four options, indicating HuggieBot 2.0 was preferred over HuggieBot 1.0 approximately 3:1. We ran several one-way analyses of variance (ANOVA) to see if participant gender, robot presentation order, or participant level of extroversion had a significant effect on which robot the user selected. No significance was found in any of these cases.\looseness-1

\subsubsection*{\textbf{Changes to Platform}}
After analyzing the results of the online study and reviewing the user feedback, we found several areas to improve our hugging robot. Some users found the initial HuggieBot 2.0 voice off-putting, so we changed the robot's voice to sound less robotic. We made the purple robe and the sweatshirt the robot's permanent outfit, as it had the highest number of votes and received many positive comments. 
We changed the color of the robot's face from its initial green to purple to match the robe and create a more polished look. Several users commented on the slow speed of HuggieBot 2.0's arms. Since the arm joints cannot move faster than the maximum angular velocity specified by the manufacturer, we instead moved their starting position closer to the goal position to reduce the time they need to close.\looseness-1

%% file: sections/06_in_person_study.tex
\section{In-Person User Study}
\label{In Person User Study}
The goal of the in-person study was to evaluate our updated robotic platform and the four new hugging tenets that drove its design. This study was also approved by the Max Planck Society ethics council under the HI framework. \looseness-1 
\subsubsection*{\textbf{Participants}}
The recruitment methods for the in-person study were the same as for the online study. Participants not employed by the Max Planck Society were compensated 12 euros. A total of 32 subjects participated in the in-person study: 37.5\% male and 62.5\% female. Our participants ranged in age from 21 to 60 (M = 30, SD = 7) and came from 13 different countries. We took significant precautions beyond government regulations to protect participant health when running this study during the COVID-19 pandemic.\looseness-1


\subsubsection*{\textbf{Procedures}}
After confirming their eligibility given the exclusion criteria, 
users scheduled an appointment for a 1.5-hour-long session with HuggieBot 2.0. Upon arrival, the experimenter explained the study, and the potential subject read the informed consent document and asked questions. If he/she still wanted to participate, the subject signed the consent form and the video release form, at which point the video cameras were turned on to record the experiment.\looseness-1 

Users began by filling out a demographics survey on a computer. 
Next, the investigator introduced the robot as the personality ``HuggieBot'' and explained its key features, including the emergency stop. The experimenter explained how the trials would work and how the subject should be prepared to move. She also explained the two different ways to initiate a hug (walking, key press) and the three different ways to be released from a hug (release hands, lean back, wait until robot releases). At this point the subject filled out an opening survey to document his/her initial impressions of the robot; participants rated on a sliding scale from 0~(disagree) to 10~(agree) how much they agreed with the statements found in Table \ref{table:OpenCloseSurvey}. Note that these questions were asked before the user had any experience physically hugging the robot. Next, the user performed practice hugs with the robot to acclimate to the hug initiation methods and the timing of the robot's arms closing. A participant was allowed to perform as many practice hugs as desired, verbally indicating to the experimenter when they were ready to begin the experiment. On average users did 2 or 3 practice hugs, but users taller than the robot (1.75~m) averaged 5 or 6 practice hugs because it took more time for them to find the most comfortable arm positions. The eight hugging conditions that made up this experiment are all three possible pairwise combinations of our three binary factors (vision, sizing, and release detection). The video associated with this paper shows a hugging trial from each of the eight conditions.


\begin{table}
\caption{The fifteen questions asked in the opening and closing questionnaires.}
\vspace{-0.4cm}
\label{table:OpenCloseSurvey}       
\begin{small}
\begin{tabularx}{\linewidth}{lX}
\hline\noalign{\smallskip}
I feel understood by the robot\\
I trust the robot\\
Robots would be nice to hug  \\
I like the presence of the robot\\
I think using the robot is a good idea \\
I am afraid to break something while using the robot\\
People would be impressed if I had such a robot\\
I could cooperate with the robot \\
I think the robot is easy to use\\
I could do activities with this robot\\
I feel threatened by the robot \\
This robot would be useful for me \\
This robot could help me\\
This robot could support me\\
I consider this robot to be a social agent\\
\noalign{\smallskip}\hline
\end{tabularx}
\end{small}
\vspace{-0.3cm}
\end{table}

We used an $8 \times 8$ Latin square to counter-balance any effects of presentation order \cite{LatinSquare} and 
recruited 32 participants to have complete Latin squares. After each hug, the participant returned to the computer and answered six questions. The first question was a free-response asking for the user's ``first impressions of this interaction.'' Then, the participant used a sliding scale from 0 to 10 to answer the five questions found in Table \ref{table:EachTrialSurvey}, which were the same questions as in the online study. A subject could request to experience the same hug again if needed. After experiencing all eight hug conditions, the participants experienced an average of 16 more hugs, during which they contacted the robot's back in different ways and received various robot responses. Data were collected for these additional hugs, but they will not be analyzed in this paper due to space constraints. At the end of the experiment, the subject answered the same questions from the beginning of the study (Table \ref{table:OpenCloseSurvey}). Finally, users could provide additional comments at the end.\looseness-1

All slider-type questions in the survey were based on previous surveys in HRI research and typical Unified Theory of Acceptance and Use of Technology (UTAUT) questionnaires \cite{UTAUT,Acceptance}. The free-response questions were designed to give the investigators any other information the participant wanted to share about the experience. A within-subjects study was selected for this experiment because we were most interested in the differences between the conditions, rather than the overall response levels to a robot hug. We also preferred this design for its higher statistical power given the same number of participants compared to a between-subjects study \cite{ExperimentalDesign}.\looseness-1

\begin{figure*}[t]
\includegraphics[trim = {6cm 0.5cm 5.25cm 1cm}, clip, width = .95\textwidth]{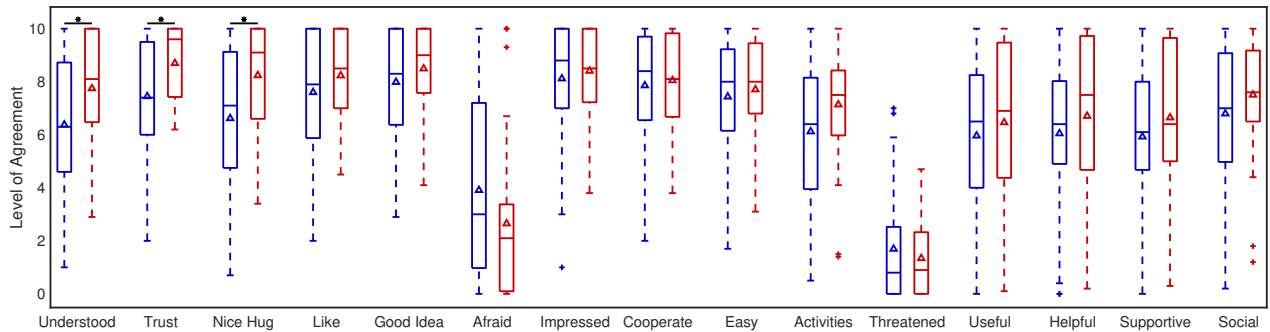}
\vspace{-0.4cm}
\caption{A comparison of the responses to the opening (blue) and closing (red) surveys. The top and bottom of the box represent the 25th and 75th percentile responses, respectively, while the line in the center represents the median, and the triangle indicates the mean. The lines extending past the boxes show the farthest data points not considered outliers. The + marks indicate outliers. The black lines with stars at the top of the graph indicate statistically significant differences.}
\vspace{-0.3cm}
\label{fig:openclose}
\end{figure*}

\subsubsection*{\textbf{Results}}
This in-person study was the first robustness test of the fully integrated HuggieBot 2.0 system. Each subject experienced a minimum of 24 hugs during the study, plus practice hugs. With 32 total participants, the robot executed more than 850 successful hugs over the course of the entire study, sometimes giving 200 hugs in one day.\looseness-1
\begin{figure*}[t]
\includegraphics[trim = {6cm 0.5cm 5.25cm 0.75cm}, clip, width = .95\textwidth]{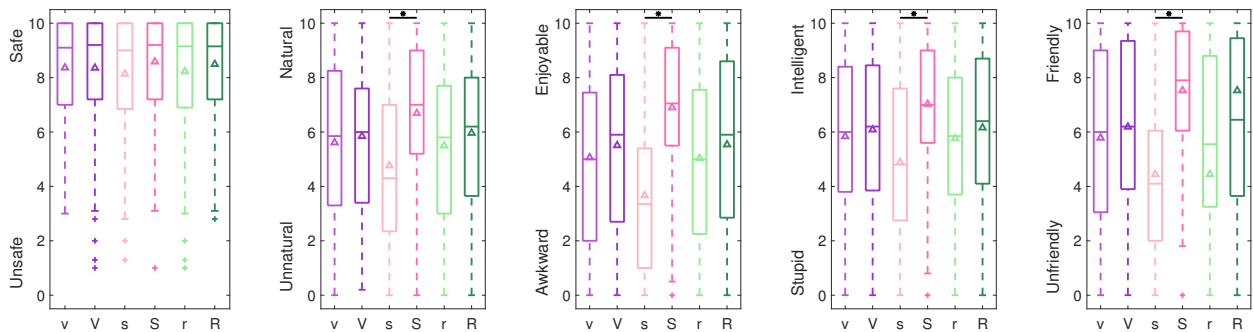}
\vspace{-0.4cm}
\caption{A comparison of the responses to the survey questions after each hug, grouped by factors. The purple colors represent without vision (v) and with vision (V), the pink colors represent without sizing (s) and with sizing (S), and the green colors represent timed release (r) and haptic release (R).}
\vspace{-0.3cm}
\label{fig:5qs}
\end{figure*}

For all statistical analyses, we applied a Bonferroni alpha correction (to account for 15 multiple comparisons) to $\alpha$~=~0.05 to determine significance. We use Pearson's linear correlation coefficient, $\rho$, to report effect size. Box plots of the responses to the opening and closing survey questions from Table~\ref{table:OpenCloseSurvey} are shown in Fig.~\ref{fig:openclose}. In this study, answers were submitted on a continuous sliding scale, so a paired t-test comparison of the opening and closing survey was conducted. We found that users felt understood by (p = 0.0025, $\rho$ = 0.57) and trusted the robot more (p < 0.001, $\rho$ = 0.70) after participating in the experiment. Users also felt that robots were nicer to hug (p < 0.001, $\rho$ = 0.76 ) 

The responses to the five questions asked after each hug can be seen grouped by the presence and absence of each of the three tested factors (vision, sizing, and release) in Fig.~\ref{fig:5qs}. These responses were analyzed using three-way repeated measures analysis of variance via the built-in MATLAB function {\tt ranova}; our data satisfy all assumptions of this analytical approach.
No significant improvements were noticed in the perceived safety of the robot in any of the tested conditions, as the robot was consistently rated highly safe. 
The automatic size adjustment significantly increased users' impressions of the naturalness of the robot's movement (F(1,31) = 25.192, p < 0.001, $\rho$ = 0.4158). 
Users found the robot's hug significantly more enjoyable when it adjusted to their size (F(1,31) = 70.553, p < 0.001, $\rho$ = 0.3610). 
Automatic size adjustment to the user caused a significant increase (F(1,31) = 25.102, p < 0.001, $\rho$ =0.4258) in the perceived intelligence of the robot. 
Finally, the robot was considered significantly friendlier when it adjusted to the size of the user (F(1,31) = 84.925, p < 0.001, $\rho$ = 0.3205). 
In summary, haptic hug sizing significantly affected every aspect except safety. 
Trials that included visual perception trended slightly positive but were not significantly different for any five of the investigated questions; small positive trends for haptic release were also not significant.\looseness-1

Eight users (25\%) verbally stated and wrote about their preference for not having to push a button to activate a hug. Out of the 256 distinct hug response surveys (32 users, 8 surveys per user), the physical warmth of the robot was positively mentioned 100 times (39\%), further validating that physical warmth is critical to pleasant robot hugs (T2) \cite{block2019softness}. These positive comments were most commonly seen in the conditions with automatic hug sizing, presumably because the increased contact with the robot torso made the heat more apparent. Additionally, we observed that our participants used a mixture of pressure release and torque release to indicate their desire to end the hugs in the study.  17 users (53\%) voiced their preference for the haptic release hugs, saying when the robot released before they were ready (in hugs with a timed release), ``he didn't want to hug me!'' or that the hug was ``too short!''  Interestingly, the hug condition where all three perceptual factors were present had the highest number of positive comments. 31 out of 32 users (96.8\%) commented that this condition was the most ``pleasant interaction,'' ``natural,'' ``friendly,'' or ``fun.''

%% file: sections/07_discussion.tex
\section{Discussion}
\label{Discussion}

Our three hypotheses were largely supported by the results. First, H1 hypothesized that when viewing from a distance, potential users will prefer the embodiment and movement of a hugging robot that incorporates our four new tenets over a state-of-the-art hugging robot that violates these tenets. 
The online study found that users significantly preferred HuggieBot 2.0 over HuggieBot 1.0; we believe our new robot was preferred because it obeys all six tenets. Written comments from the online community mention the ``large,'' ``hulking,'' and ``over-powering'' PR2 robot as unnerving when compared to the size of the user. In comparison, our robot is considered ``nice'' and ``friendlier.'' Several users also wrote comments on how the people in the PR2 videos had to push a button on the robot's back, which seemed ``unnatural'', whereas the HuggieBot 2.0 release seemed more ``intuitive''. We did our best to match the videos of our new robot to the pre-existing videos of the PR2 so as not to bias the online viewers. These videos included but could not showcase visual hug initiation and haptic size adjustment. 
Based on the strong preference for our hugging robot in our carefully controlled online study, we conclude that users do prefer the embodiment and movement of a hugging robot that obeys our four new tenets over a state-of-the-art hugging robot that violates most of them. \looseness-1

H2 conjectured that obeying the four new tenets during physical interactions with users will significantly increase the perceived safety, naturalness, enjoyability, intelligence, and friendliness of a hugging robot. We found that the haptic perception tenets had the greatest effects on these aspects of the robot, with haptic sizing positively affecting many responses and both haptic sizing and haptic release garnering positive comments. 
The lack of significant effects of visual initiation does not match the comments that users prefer the interaction when the robot recognizes their approach, rather than them having to push a button to initiate the hug. It is possible that users might not have included the button pushing in their rating of the hug as we simply asked users to ``rate their experience with this hug'' and did not explicitly tell them to include the hug initiation. Users might also have been confused that they had to walk towards the robot to initiate a hug, and then the robot would ask ``Can I have a hug, please?'' We chose to have the robot say the same phrase for both initiation methods to minimize variables, but as the user was initiating the hug, it may have made more sense for the robot to say something else or not speak. \looseness-1

We also believe there is room for improvement of the visual perception of our robot, which could contribute to higher ratings of the five questions. Currently, our perception of the user is based solely on his/her approach. To take perception even further, we believe hugs would be more comfortable if the robot could adjust its arm poses to match the approaching user's height and arm positions. Our taller users found the robot hugged them too low, and our shorter users found the opposite. 
Adjusting to user height would more fully obey T4 (visual perception) and therefore should be more acceptable to users. While the torso of the robot and dual release methods ensure our robot follows T6 of reliably releasing the user, a robot that could adjust its arm positions to the reciprocating pose of the user could greatly strengthen the user's impression of the robot's visual perception and improve the user opinion of the robot. We concede that our robot's rudimentary visual perception contributed to our lack of finding significant differences in the areas we investigated when testing with and without this factor. \looseness-1

Finally, H3 hypothesized that repeated hugs with a robot that follows all six tenets will improve user opinions about robots in general. We asked the same opening and closing survey questions as \citet{block2019softness}, with similar results. 
Both studies' users felt more understood by, trusted, and thought that robots were nicer to hug after participating. HuggieBot 1.0's users also liked the presence of the robot more afterwards, found the robot easier to use, and viewed it as more of a social agent after the experiment, although these findings were reported without any statistical correction for multiple comparisons. The PR2 robot used in that experiment is significantly larger than an adult human, which violates T3. This domineering physical presence, therefore, contributed to lower initial ratings for users liking the presence of the robot, their perceived ease of use of the robot, and viewing the robot as a social agent. Our new robot, whose physical stature obeys the first three tenets, received higher initial ratings in these categories. HuggieBot 2.0 appeared as a friendly social agent from the beginning, and prolonged interaction with it confirmed these high initial impressions, which is why we did not find any significant differences for these questions in our study. As first impressions are often critical to determine whether a user will interact with a robot, here we see that it is important to obey the tenet prescribing the physical size of a robot. Therefore, we conclude that a robot that follows the six tenets does indeed improve user opinions about robots in general. A positive impression of the robot is crucial because it will make users more willing to receive a robot hug, and thus more likely to receive these health benefits when they cannot receive them from other people.  \looseness-1


%% file: sections/08_conclusion.tex
\section{Limitations and Conclusion}
\label{Conclusion}

This study represents an important step in understanding intimate social-physical human-robot interactions, but it certainly has limitations. Due to COVID-19, the first study relied solely on videos and images, rather than the participants physically interacting with robots. Since we do not have access to a PR2, this online study enabled a fair comparison between our new platform and a previously well-rated hugging robot. Watching other people hug a robot is how users will decide if they also want to interact with a hugging robot in the wild.  

One weakness of our new platform, HuggieBot 2.0, was the slow speed of the Kinova JACO arms. We selected these arms because of their inherent safety features; however, the distance the arms had to travel made their slow speed obvious and caused a long delay after hug initiation. When users started the hug with a button press, they could wait before walking to the robot for better timing. With visual hug initiation, the users were required to walk before the arms began moving, which resulted in many awkwardly waiting in front of the robot for the arms to close. 
Related to this limitation is the size of the room in which we conducted the experiment. A larger room would have let us set the threshold distance farther back to accommodate the speed of the arms. 
Both of these limitations could have contributed to the lack of significant differences between the hugs with and without visual perception. \looseness-1

Another limitation is the self-selection bias of our participants. For transparency, we advertised the experiment as a hugging robot study. While we succeeded in recruiting a diverse and largely non-technical audience to make our results as applicable to the general public as possible, we nevertheless acknowledge that users who chose to participate in the study were interested in robots. Because we did not hide the nature of our study, we did not have any participants who refused to hug the robot, as might occur in a more natural in-the-wild study design.\looseness-1

This project took a critical look at state-of-the-art hugging robots, improved upon their flaws, built upon their successes, and created a new hugging robot, HuggieBot 2.0. We also propose to the HRI community six tenets of hugging that future designers should consider to improve user acceptance of hugging robots. During times of social distancing, the consequences of lack of physical contact with others can be more damaging and prevalent than ever. If we cannot seek comfort from other people due to physical distance or health or safety concerns, it is important that we seek other opportunities to reap the benefits of this helpful interaction. \looseness-1

%% file: sections/appendix.tex








%% file: Block_HRI_2021.bbl

\begin{thebibliography}{44}


\ifx \showCODEN    \undefined \def \showCODEN     #1{\unskip}     \fi
\ifx \showDOI      \undefined \def \showDOI       #1{#1}\fi
\ifx \showISBNx    \undefined \def \showISBNx     #1{\unskip}     \fi
\ifx \showISBNxiii \undefined \def \showISBNxiii  #1{\unskip}     \fi
\ifx \showISSN     \undefined \def \showISSN      #1{\unskip}     \fi
\ifx \showLCCN     \undefined \def \showLCCN      #1{\unskip}     \fi
\ifx \shownote     \undefined \def \shownote      #1{#1}          \fi
\ifx \showarticletitle \undefined \def \showarticletitle #1{#1}   \fi
\ifx \showURL      \undefined \def \showURL       {\relax}        \fi
\providecommand\bibfield[2]{#2}
\providecommand\bibinfo[2]{#2}
\providecommand\natexlab[1]{#1}
\providecommand\showeprint[2][]{arXiv:#2}

\bibitem[\protect\citeauthoryear{Barber, Volz, Desai, Rubinfeld, Schipper, and
  Wolter}{Barber et~al\mbox{.}}{1986}]%
        {GripSelection}
\bibfield{author}{\bibinfo{person}{James Barber}, \bibinfo{person}{Richard~A.
  Volz}, \bibinfo{person}{Rajiv Desai}, \bibinfo{person}{Ronitt Rubinfeld},
  \bibinfo{person}{Brian Schipper}, {and} \bibinfo{person}{Jan Wolter}.}
  \bibinfo{year}{1986}\natexlab{}.
\newblock \showarticletitle{Automatic two-fingered grip selection}. In
  \bibinfo{booktitle}{\emph{Proceedings of the IEEE International Conference on
  Robotics and Automation (ICRA)}}, Vol.~\bibinfo{volume}{3}.
  \bibinfo{pages}{890--896}.
\newblock


\bibitem[\protect\citeauthoryear{Block and Kuchenbecker}{Block and
  Kuchenbecker}{2018}]%
        {block2018emotionally}
\bibfield{author}{\bibinfo{person}{Alexis~E. Block} {and}
  \bibinfo{person}{Katherine~J. Kuchenbecker}.}
  \bibinfo{year}{2018}\natexlab{}.
\newblock \showarticletitle{Emotionally supporting humans through robot hugs}.
  In \bibinfo{booktitle}{\emph{Companion of the ACM/IEEE International
  Conference on Human-Robot Interaction}}. \bibinfo{address}{Chicago, USA},
  \bibinfo{pages}{293--294}.
\newblock


\bibitem[\protect\citeauthoryear{Block and Kuchenbecker}{Block and
  Kuchenbecker}{2019}]%
        {block2019softness}
\bibfield{author}{\bibinfo{person}{Alexis~E. Block} {and}
  \bibinfo{person}{Katherine~J. Kuchenbecker}.}
  \bibinfo{year}{2019}\natexlab{}.
\newblock \showarticletitle{Softness, warmth, and responsiveness improve robot
  hugs}.
\newblock \bibinfo{journal}{\emph{International Journal of Social Robotics}}
  \bibinfo{volume}{11}, \bibinfo{number}{1} (\bibinfo{year}{2019}),
  \bibinfo{pages}{49--64}.
\newblock
\urldef\tempurl%
\url{https://doi.org/10.1007/s12369-018-0495-2}
\showDOI{\tempurl}


\bibitem[\protect\citeauthoryear{Cascio, Moore, and McGlone}{Cascio
  et~al\mbox{.}}{2019}]%
        {SocialTouchDevelopment}
\bibfield{author}{\bibinfo{person}{Carissa~J. Cascio}, \bibinfo{person}{David
  Moore}, {and} \bibinfo{person}{Francis McGlone}.}
  \bibinfo{year}{2019}\natexlab{}.
\newblock \showarticletitle{Social touch and human development}.
\newblock \bibinfo{journal}{\emph{Developmental Cognitive Neuroscience}}
  \bibinfo{volume}{35} (\bibinfo{year}{2019}), \bibinfo{pages}{5--11}.
\newblock
\showISSN{1878-9293}
\urldef\tempurl%
\url{https://doi.org/10.1016/j.dcn.2018.04.009}
\showDOI{\tempurl}


\bibitem[\protect\citeauthoryear{{Choi}, {Pantofaru}, and {Savarese}}{{Choi}
  et~al\mbox{.}}{2011}]%
        {choi2011}
\bibfield{author}{\bibinfo{person}{Wongun {Choi}}, \bibinfo{person}{Caroline
  {Pantofaru}}, {and} \bibinfo{person}{Silvio {Savarese}}.}
  \bibinfo{year}{2011}\natexlab{}.
\newblock \showarticletitle{Detecting and tracking people using an {RGB-D}
  camera via multiple detector fusion}. In
  \bibinfo{booktitle}{\emph{Proceedings of the IEEE International Conference on
  Computer Vision Workshops (ICCV Workshops)}}. \bibinfo{pages}{1076--1083}.
\newblock


\bibitem[\protect\citeauthoryear{Cohen, Janicki-Deverts, Turner, and
  Doyle}{Cohen et~al\mbox{.}}{2015}]%
        {HuggingHealthBenefits}
\bibfield{author}{\bibinfo{person}{Sheldon Cohen}, \bibinfo{person}{Denise
  Janicki-Deverts}, \bibinfo{person}{Ronald~B. Turner}, {and}
  \bibinfo{person}{William~J. Doyle}.} \bibinfo{year}{2015}\natexlab{}.
\newblock \showarticletitle{Does hugging provide stress-buffering social
  support? a study of susceptibility to upper respiratory infection and
  illness}.
\newblock \bibinfo{journal}{\emph{Psychological Science}} \bibinfo{volume}{26},
  \bibinfo{number}{2} (\bibinfo{year}{2015}), \bibinfo{pages}{135--147}.
\newblock
\urldef\tempurl%
\url{https://doi.org/10.1177/0956797614559284}
\showDOI{\tempurl}


\bibitem[\protect\citeauthoryear{{Costanzo}, {De Maria}, and
  {Natale}}{{Costanzo} et~al\mbox{.}}{2020}]%
        {TwoFingeredGrasp}
\bibfield{author}{\bibinfo{person}{Marco {Costanzo}}, \bibinfo{person}{Giuseppe
  {De Maria}}, {and} \bibinfo{person}{Ciro {Natale}}.}
  \bibinfo{year}{2020}\natexlab{}.
\newblock \showarticletitle{Two-fingered in-hand object handling based on
  force/tactile feedback}.
\newblock \bibinfo{journal}{\emph{IEEE Transactions on Robotics}}
  \bibinfo{volume}{36}, \bibinfo{number}{1} (\bibinfo{year}{2020}),
  \bibinfo{pages}{157--173}.
\newblock


\bibitem[\protect\citeauthoryear{Dalal, Triggs, and Schmid}{Dalal
  et~al\mbox{.}}{2006}]%
        {dalal2006}
\bibfield{author}{\bibinfo{person}{Navneet Dalal}, \bibinfo{person}{Bill
  Triggs}, {and} \bibinfo{person}{Cordelia Schmid}.}
  \bibinfo{year}{2006}\natexlab{}.
\newblock \showarticletitle{Human detection using oriented histograms of flow
  and appearance}. In \bibinfo{booktitle}{\emph{Proceedings of the European
  Conference on Computer Vision (ECCV) - Volume Part II}} (Graz, Austria).
  \bibinfo{publisher}{Springer-Verlag}, \bibinfo{address}{Berlin, Heidelberg},
  \bibinfo{pages}{428--441}.
\newblock
\showISBNx{3540338349}
\urldef\tempurl%
\url{https://doi.org/10.1007/11744047_33}
\showDOI{\tempurl}


\bibitem[\protect\citeauthoryear{de~Winter and Dodou}{de~Winter and
  Dodou}{2017}]%
        {ExperimentalDesign}
\bibfield{author}{\bibinfo{person}{Joost C.~F. de Winter} {and}
  \bibinfo{person}{Dimitra Dodou}.} \bibinfo{year}{2017}\natexlab{}.
\newblock \showarticletitle{Experimental design in: human subject research for
  engineers}.
\newblock \bibinfo{journal}{\emph{SpringerBriefs in Applied Sciences and
  Technology}} (\bibinfo{year}{2017}).
\newblock


\bibitem[\protect\citeauthoryear{DiSalvo, Gemperle, Forlizzi, and
  Montgomery}{DiSalvo et~al\mbox{.}}{2003}]%
        {disalvo2003hug}
\bibfield{author}{\bibinfo{person}{Carl DiSalvo}, \bibinfo{person}{Francine
  Gemperle}, \bibinfo{person}{Jodi Forlizzi}, {and} \bibinfo{person}{Elliott
  Montgomery}.} \bibinfo{year}{2003}\natexlab{}.
\newblock \showarticletitle{The {H}ug: an exploration of robotic form for
  intimate communication}. In \bibinfo{booktitle}{\emph{Proceedings of the IEEE
  International Symposium on Robot and Human Interactive Communication
  (RO-MAN)}}. \bibinfo{pages}{403--408}.
\newblock


\bibitem[\protect\citeauthoryear{Duvall, Dunne, Schleif, and Holschuh}{Duvall
  et~al\mbox{.}}{2016}]%
        {duvall2016active}
\bibfield{author}{\bibinfo{person}{Julia~C. Duvall}, \bibinfo{person}{Lucy~E.
  Dunne}, \bibinfo{person}{Nicholas Schleif}, {and} \bibinfo{person}{Brad
  Holschuh}.} \bibinfo{year}{2016}\natexlab{}.
\newblock \showarticletitle{Active hugging vest for deep touch pressure
  therapy}. In \bibinfo{booktitle}{\emph{Proceedings of the ACM International
  Joint Conference on Pervasive and Ubiquitous Computing: Adjunct}}.
  \bibinfo{pages}{458--463}.
\newblock


\bibitem[\protect\citeauthoryear{Edelson, Goldberg, Kerr, and Grandin}{Edelson
  et~al\mbox{.}}{1999}]%
        {Edelson1999}
\bibfield{author}{\bibinfo{person}{Stephen~M. Edelson},
  \bibinfo{person}{Meredyth Goldberg}, \bibinfo{person}{David C.~R. Kerr},
  {and} \bibinfo{person}{Temple Grandin}.} \bibinfo{year}{1999}\natexlab{}.
\newblock \showarticletitle{Behavioral and Physiological Effects of Deep
  Pressure on Children With Autism: A Pilot Study Evaluating the Efficacy of
  Grandin's Hug Machine}.
\newblock  \bibinfo{volume}{53}, \bibinfo{number}{1979} (\bibinfo{year}{1999}),
  \bibinfo{pages}{145--152}.
\newblock


\bibitem[\protect\citeauthoryear{Fitter and Kuchenbecker}{Fitter and
  Kuchenbecker}{2016}]%
        {fitter2016designing}
\bibfield{author}{\bibinfo{person}{Naomi~T. Fitter} {and}
  \bibinfo{person}{Katherine~J. Kuchenbecker}.}
  \bibinfo{year}{2016}\natexlab{}.
\newblock \showarticletitle{Designing and assessing expressive open-source
  faces for the Baxter robot}. In \bibinfo{booktitle}{\emph{Proceedings of the
  International Conference on Social Robotics}}. Springer,
  \bibinfo{pages}{340--350}.
\newblock


\bibitem[\protect\citeauthoryear{Grant}{Grant}{1948}]%
        {LatinSquare}
\bibfield{author}{\bibinfo{person}{David~A. Grant}.}
  \bibinfo{year}{1948}\natexlab{}.
\newblock \showarticletitle{The {L}atin square principle in the design and
  analysis of psychological experiments.}
\newblock \bibinfo{journal}{\emph{Psychological Bulletin}}
  \bibinfo{volume}{45}, \bibinfo{number}{5} (\bibinfo{year}{1948}),
  \bibinfo{pages}{427--442}.
\newblock


\bibitem[\protect\citeauthoryear{Hall, Birdwhistell, Bock, Bohannan, Jr.,
  Durbin, Edmonson, Fischer, Hymes, Kimball, Barre, Lynch, J, McClellan,
  Marshall, Milner, Sarles, Trager, and Vayda}{Hall et~al\mbox{.}}{1968}]%
        {Proxemics}
\bibfield{author}{\bibinfo{person}{Edward~T. Hall}, \bibinfo{person}{Ray~L.
  Birdwhistell}, \bibinfo{person}{Bernhard Bock}, \bibinfo{person}{Paul
  Bohannan}, \bibinfo{person}{A.~Richard~Diebold Jr.},
  \bibinfo{person}{Marshall Durbin}, \bibinfo{person}{Munro~S. Edmonson},
  \bibinfo{person}{J.~L. Fischer}, \bibinfo{person}{Dell Hymes},
  \bibinfo{person}{Solon~T. Kimball}, \bibinfo{person}{Weston~La Barre},
  \bibinfo{person}{Franck Lynch}, \bibinfo{person}{S. J},
  \bibinfo{person}{J.~E. McClellan}, \bibinfo{person}{Donald~S. Marshall},
  \bibinfo{person}{G.~B. Milner}, \bibinfo{person}{Harvey~B. Sarles},
  \bibinfo{person}{George~L. Trager}, {and} \bibinfo{person}{Andrew~P. Vayda}.}
  \bibinfo{year}{1968}\natexlab{}.
\newblock \showarticletitle{Proxemics [and {C}omments and {R}eplies]}.
\newblock \bibinfo{journal}{\emph{Current Anthropology}} \bibinfo{volume}{9},
  \bibinfo{number}{2/3} (\bibinfo{year}{1968}), \bibinfo{pages}{83--108}.
\newblock


\bibitem[\protect\citeauthoryear{Hedayati, Bhaduri, Sumner, Szafir, and
  Gross}{Hedayati et~al\mbox{.}}{2019}]%
        {hedayati2019hugbot}
\bibfield{author}{\bibinfo{person}{Hooman Hedayati}, \bibinfo{person}{Srinjita
  Bhaduri}, \bibinfo{person}{Tamara Sumner}, \bibinfo{person}{Daniel Szafir},
  {and} \bibinfo{person}{Mark~D. Gross}.} \bibinfo{year}{2019}\natexlab{}.
\newblock \showarticletitle{{HugBot}: a soft robot designed to give human-like
  hugs}. In \bibinfo{booktitle}{\emph{Proceedings of the ACM International
  Conference on Interaction Design and Children}}. \bibinfo{pages}{556--561}.
\newblock


\bibitem[\protect\citeauthoryear{Heerink, Krose, Evers, and Wielinga}{Heerink
  et~al\mbox{.}}{2009}]%
        {Acceptance}
\bibfield{author}{\bibinfo{person}{Marcel Heerink}, \bibinfo{person}{Ben
  Krose}, \bibinfo{person}{Vanessa Evers}, {and} \bibinfo{person}{Bob~J.
  Wielinga}.} \bibinfo{year}{2009}\natexlab{}.
\newblock \showarticletitle{Measuring acceptance of an assistive social robot:
  a suggested toolkit}. In \bibinfo{booktitle}{\emph{Proceedings of the IEEE
  International Symposium on Robot and Human Interactive Communication
  (RO-MAN)}}. \bibinfo{pages}{528--533}.
\newblock


\bibitem[\protect\citeauthoryear{Liu, Anguelov, Erhan, Szegedy, Reed, Fu, and
  Berg}{Liu et~al\mbox{.}}{2016}]%
        {Liu_2016}
\bibfield{author}{\bibinfo{person}{Wei Liu}, \bibinfo{person}{Dragomir
  Anguelov}, \bibinfo{person}{Dumitru Erhan}, \bibinfo{person}{Christian
  Szegedy}, \bibinfo{person}{Scott Reed}, \bibinfo{person}{Cheng-Yang Fu},
  {and} \bibinfo{person}{Alexander~C. Berg}.} \bibinfo{year}{2016}\natexlab{}.
\newblock \showarticletitle{SSD: Single Shot MultiBox Detector}.
\newblock \bibinfo{journal}{\emph{Lecture Notes in Computer Science}}
  (\bibinfo{year}{2016}), \bibinfo{pages}{21--37}.
\newblock
\showISBNx{9783319464480}
\showISSN{1611-3349}
\urldef\tempurl%
\url{https://doi.org/10.1007/978-3-319-46448-0_2}
\showDOI{\tempurl}


\bibitem[\protect\citeauthoryear{Ma, Spiers, and Dollar}{Ma
  et~al\mbox{.}}{2016}]%
        {SpiersGrasping}
\bibfield{author}{\bibinfo{person}{Raymond~R. Ma}, \bibinfo{person}{Adam
  Spiers}, {and} \bibinfo{person}{Aaron~M. Dollar}.}
  \bibinfo{year}{2016}\natexlab{}.
\newblock \showarticletitle{{M2} gripper: extending the dexterity of a simple,
  underactuated gripper}.
\newblock \bibinfo{journal}{\emph{Advances in Reconfigurable Mechanisms and
  Robots II. Mechanisms and Machine Science}}  \bibinfo{volume}{36}
  (\bibinfo{year}{2016}).
\newblock
\urldef\tempurl%
\url{https://doi.org/10.1007/978-3-319-23327-7_68}
\showDOI{\tempurl}


\bibitem[\protect\citeauthoryear{Miyashita and Ishiguro}{Miyashita and
  Ishiguro}{2004}]%
        {miyashita2004human}
\bibfield{author}{\bibinfo{person}{Takahiro Miyashita} {and}
  \bibinfo{person}{Hiroshi Ishiguro}.} \bibinfo{year}{2004}\natexlab{}.
\newblock \showarticletitle{Human-like natural behavior generation based on
  involuntary motions for humanoid robots}.
\newblock \bibinfo{journal}{\emph{Robotics and Autonomous Systems}}
  \bibinfo{volume}{48}, \bibinfo{number}{4} (\bibinfo{year}{2004}),
  \bibinfo{pages}{203--212}.
\newblock


\bibitem[\protect\citeauthoryear{Morrison and Gore}{Morrison and Gore}{2010}]%
        {InternetAndDepression}
\bibfield{author}{\bibinfo{person}{Catriona~M. Morrison} {and}
  \bibinfo{person}{Helen Gore}.} \bibinfo{year}{2010}\natexlab{}.
\newblock \showarticletitle{The relationship between excessive {I}nternet use
  and depression: a questionnaire-based study of 1,319 young people and
  adults}.
\newblock \bibinfo{journal}{\emph{Psychopathology}} \bibinfo{volume}{43},
  \bibinfo{number}{2} (\bibinfo{year}{2010}), \bibinfo{pages}{121--126}.
\newblock
\showISBNx{0254-4962}
\urldef\tempurl%
\url{https://doi.org/10.1159/000277001}
\showDOI{\tempurl}


\bibitem[\protect\citeauthoryear{Neira and Barber}{Neira and Barber}{2014}]%
        {DepressedTeensFromSocialMedia}
\bibfield{author}{\bibinfo{person}{Corey J.~Blomfield Neira} {and}
  \bibinfo{person}{Bonnie~L. Barber}.} \bibinfo{year}{2014}\natexlab{}.
\newblock \showarticletitle{Social networking site use: linked to adolescents'
  social self-concept, self-esteem, and depressed mood}.
\newblock \bibinfo{journal}{\emph{Australian Journal of Psychology}}
  \bibinfo{volume}{66}, \bibinfo{number}{1} (\bibinfo{year}{2014}),
  \bibinfo{pages}{56--64}.
\newblock
\urldef\tempurl%
\url{https://doi.org/10.1111/ajpy.12034}
\showDOI{\tempurl}


\bibitem[\protect\citeauthoryear{Odabasi}{Odabasi}{2017}]%
        {Odabasi2017}
\bibfield{author}{\bibinfo{person}{Cagatay Odabasi}.}
  \bibinfo{year}{2017}\natexlab{}.
\newblock \bibinfo{title}{{ROS} people object detection and action recognition
  {T}ensorflow}.
\newblock
\newblock
\urldef\tempurl%
\url{https://github.com/cagbal/ros_people_object_detection_tensorflow}
\showURL{%
\tempurl}


\bibitem[\protect\citeauthoryear{{Oren}, {Papageorgiou}, {Sinha}, {Osuna}, and
  {Poggio}}{{Oren} et~al\mbox{.}}{1997}]%
        {wavelet1997}
\bibfield{author}{\bibinfo{person}{Mike {Oren}}, \bibinfo{person}{Constantine
  {Papageorgiou}}, \bibinfo{person}{P. {Sinha}}, \bibinfo{person}{E. {Osuna}},
  {and} \bibinfo{person}{Tomaso {Poggio}}.} \bibinfo{year}{1997}\natexlab{}.
\newblock \showarticletitle{Pedestrian detection using wavelet templates}. In
  \bibinfo{booktitle}{\emph{Proceedings of IEEE Computer Society Conference on
  Computer Vision and Pattern Recognition}}. \bibinfo{pages}{193--199}.
\newblock


\bibitem[\protect\citeauthoryear{Paul, Haque, and Chakraborty}{Paul
  et~al\mbox{.}}{2013}]%
        {manoranjan2013surveillance}
\bibfield{author}{\bibinfo{person}{Manoranjan Paul}, \bibinfo{person}{Shah
  Haque}, {and} \bibinfo{person}{Subrata Chakraborty}.}
  \bibinfo{year}{2013}\natexlab{}.
\newblock \showarticletitle{Human detection in surveillance videos and its
  applications: a review}.
\newblock \bibinfo{journal}{\emph{Eurasip Journal on Applied Signal
  Processing}}  \bibinfo{volume}{176} (\bibinfo{year}{2013}),
  \bibinfo{pages}{1--16}.
\newblock
\showISSN{0941-0635}
\urldef\tempurl%
\url{https://doi.org/10.1186/1687-6180-2013-176}
\showDOI{\tempurl}


\bibitem[\protect\citeauthoryear{Romano, Hsiao, Niemeyer, Chitta, and
  Kuchenbecker}{Romano et~al\mbox{.}}{2011}]%
        {Romano11-TRO-Grasp}
\bibfield{author}{\bibinfo{person}{Joseph~M. Romano}, \bibinfo{person}{Kaijen
  Hsiao}, \bibinfo{person}{G{\"u}nter Niemeyer}, \bibinfo{person}{Sachin
  Chitta}, {and} \bibinfo{person}{Katherine~J. Kuchenbecker}.}
  \bibinfo{year}{2011}\natexlab{}.
\newblock \showarticletitle{Human-inspired robotic grasp control with tactile
  sensing}.
\newblock \bibinfo{journal}{\emph{IEEE Transactions on Robotics}}
  \bibinfo{volume}{27}, \bibinfo{number}{6} (\bibinfo{date}{December}
  \bibinfo{year}{2011}), \bibinfo{pages}{1067--1079}.
\newblock


\bibitem[\protect\citeauthoryear{Scott}{Scott}{2020}]%
        {bestthing}
\bibfield{author}{\bibinfo{person}{Tom Scott}.}
  \bibinfo{year}{2020}\natexlab{}.
\newblock \showarticletitle{1,204,986 Votes Decided: What Is The Best Thing?}
\newblock \bibinfo{journal}{\emph{YouTube}} (\bibinfo{year}{2020}).
\newblock
\urldef\tempurl%
\url{https://www.youtube.com/watch?v=ALy6e7GbDR}
\showURL{%
\tempurl}


\bibitem[\protect\citeauthoryear{Shiomi, Nakata, Kanbara, and Hagita}{Shiomi
  et~al\mbox{.}}{2017a}]%
        {shiomi2017hug}
\bibfield{author}{\bibinfo{person}{Masahiro Shiomi}, \bibinfo{person}{Aya
  Nakata}, \bibinfo{person}{Masayuki Kanbara}, {and} \bibinfo{person}{Norihiro
  Hagita}.} \bibinfo{year}{2017}\natexlab{a}.
\newblock \showarticletitle{A hug from a robot encourages prosocial behavior}.
  In \bibinfo{booktitle}{\emph{Proceedings of the IEEE International Symposium
  on Robot and Human Interactive Communication (RO-MAN)}}.
  \bibinfo{pages}{418--423}.
\newblock


\bibitem[\protect\citeauthoryear{Shiomi, Nakata, Kanbara, and Hagita}{Shiomi
  et~al\mbox{.}}{2017b}]%
        {shiomi2017robot}
\bibfield{author}{\bibinfo{person}{Masahiro Shiomi}, \bibinfo{person}{Aya
  Nakata}, \bibinfo{person}{Masayuki Kanbara}, {and} \bibinfo{person}{Norihiro
  Hagita}.} \bibinfo{year}{2017}\natexlab{b}.
\newblock \showarticletitle{A robot that encourages self-disclosure by hug}. In
  \bibinfo{booktitle}{\emph{Proceedings of the International Conference on
  Social Robotics (ICSR)}}. Springer, \bibinfo{pages}{324--333}.
\newblock


\bibitem[\protect\citeauthoryear{Stiehl, Lieberman, Breazeal, Basel, Lalla, and
  Wolf}{Stiehl et~al\mbox{.}}{2005}]%
        {TheHuggable}
\bibfield{author}{\bibinfo{person}{Walter~Dan Stiehl}, \bibinfo{person}{Jeff
  Lieberman}, \bibinfo{person}{Cynthia Breazeal}, \bibinfo{person}{Louis
  Basel}, \bibinfo{person}{Levi Lalla}, {and} \bibinfo{person}{Michael Wolf}.}
  \bibinfo{year}{2005}\natexlab{}.
\newblock \showarticletitle{Design of a therapeutic robotic companion for
  relational, affective touch}. In \bibinfo{booktitle}{\emph{Proceedings of the
  IEEE International Symposium on Robot and Human Interactive Communication
  (RO-MAN)}}. \bibinfo{pages}{408--415}.
\newblock


\bibitem[\protect\citeauthoryear{Sumioka, Nakae, Kanai, and Ishiguro}{Sumioka
  et~al\mbox{.}}{2013}]%
        {HuggablePillowWithPhone}
\bibfield{author}{\bibinfo{person}{Hidenobu Sumioka}, \bibinfo{person}{Aya
  Nakae}, \bibinfo{person}{Ryota Kanai}, {and} \bibinfo{person}{Hiroshi
  Ishiguro}.} \bibinfo{year}{2013}\natexlab{}.
\newblock \showarticletitle{Huggable communication medium decreases cortisol
  levels}.
\newblock \bibinfo{journal}{\emph{Scientific Reports}} \bibinfo{volume}{3},
  \bibinfo{number}{1} (\bibinfo{year}{2013}), \bibinfo{pages}{3034}.
\newblock
\showISBNx{2045-2322}
\urldef\tempurl%
\url{https://doi.org/10.1038/srep03034}
\showDOI{\tempurl}


\bibitem[\protect\citeauthoryear{Suvilehto, Glerean, Dunbar, Hari, and
  Nummenmaa}{Suvilehto et~al\mbox{.}}{2015}]%
        {TopographyOfSocialTouch}
\bibfield{author}{\bibinfo{person}{Juulia~T. Suvilehto},
  \bibinfo{person}{Enrico Glerean}, \bibinfo{person}{Robin I.~M. Dunbar},
  \bibinfo{person}{Riitta Hari}, {and} \bibinfo{person}{Lauri Nummenmaa}.}
  \bibinfo{year}{2015}\natexlab{}.
\newblock \showarticletitle{Topography of social touching depends on emotional
  bonds between humans}.
\newblock \bibinfo{journal}{\emph{Proceedings of the National Academy of
  Sciences}} \bibinfo{volume}{112}, \bibinfo{number}{45}
  (\bibinfo{year}{2015}), \bibinfo{pages}{13811--13816}.
\newblock
\showISSN{0027-8424}
\urldef\tempurl%
\url{https://doi.org/10.1073/pnas.1519231112}
\showDOI{\tempurl}


\bibitem[\protect\citeauthoryear{Teh, Cheok, Peiris, Choi, Thuong, and Lai}{Teh
  et~al\mbox{.}}{2008}]%
        {teh2008huggypajama}
\bibfield{author}{\bibinfo{person}{James Keng~Soon Teh},
  \bibinfo{person}{Adrian~David Cheok}, \bibinfo{person}{Roshan~L. Peiris},
  \bibinfo{person}{Yongsoon Choi}, \bibinfo{person}{Vuong Thuong}, {and}
  \bibinfo{person}{Sha Lai}.} \bibinfo{year}{2008}\natexlab{}.
\newblock \showarticletitle{{H}uggy {P}ajama: a mobile parent and child hugging
  communication system}. In \bibinfo{booktitle}{\emph{Proceedings of the ACM
  International Conference on Interaction Design and Children}}.
  \bibinfo{pages}{250--257}.
\newblock


\bibitem[\protect\citeauthoryear{Townsend}{Townsend}{2000}]%
        {BarrettHand}
\bibfield{author}{\bibinfo{person}{William Townsend}.}
  \bibinfo{year}{2000}\natexlab{}.
\newblock \showarticletitle{The {BarrettHand} grasper: a programmably flexible
  part handling and assembly}.
\newblock \bibinfo{journal}{\emph{Industrial Robot: An International Journal}}
  \bibinfo{volume}{27}, \bibinfo{number}{3} (\bibinfo{year}{2000}),
  \bibinfo{pages}{181--188}.
\newblock


\bibitem[\protect\citeauthoryear{Trovato, Do, Terlemez, Mandery, Ishii,
  Bianchi-Berthouze, Asfour, and Takanishi}{Trovato et~al\mbox{.}}{2016}]%
        {trovato2016hugging}
\bibfield{author}{\bibinfo{person}{Gabriele Trovato}, \bibinfo{person}{Martin
  Do}, \bibinfo{person}{{\"O}mer Terlemez}, \bibinfo{person}{Christian
  Mandery}, \bibinfo{person}{Hiroyuki Ishii}, \bibinfo{person}{Nadia
  Bianchi-Berthouze}, \bibinfo{person}{Tamim Asfour}, {and}
  \bibinfo{person}{Atsuo Takanishi}.} \bibinfo{year}{2016}\natexlab{}.
\newblock \showarticletitle{Is hugging a robot weird? Investigating the
  influence of robot appearance on users' perception of hugging}. In
  \bibinfo{booktitle}{\emph{Proceedings of the IEEE-RAS International
  Conference on Humanoid Robots (Humanoids)}}. \bibinfo{pages}{318--323}.
\newblock


\bibitem[\protect\citeauthoryear{Tsetserukou}{Tsetserukou}{2010}]%
        {tsetserukou2010haptihug}
\bibfield{author}{\bibinfo{person}{Dzmitry Tsetserukou}.}
  \bibinfo{year}{2010}\natexlab{}.
\newblock \showarticletitle{{HaptiHug}: a novel haptic display for
  communication of hug over a distance}. In
  \bibinfo{booktitle}{\emph{Proceedings of the International Conference on
  Human Haptic Sensing and Touch Enabled Computer Applications}}. Springer,
  \bibinfo{pages}{340--347}.
\newblock


\bibitem[\protect\citeauthoryear{{Tuzel}, {Porikli}, and {Meer}}{{Tuzel}
  et~al\mbox{.}}{2007}]%
        {tuzel2007}
\bibfield{author}{\bibinfo{person}{Oncel {Tuzel}}, \bibinfo{person}{Fatih
  {Porikli}}, {and} \bibinfo{person}{Peter {Meer}}.}
  \bibinfo{year}{2007}\natexlab{}.
\newblock \showarticletitle{Human detection via classification on {R}iemannian
  manifolds}. In \bibinfo{booktitle}{\emph{Proceedings of the IEEE Conference
  on Computer Vision and Pattern Recognition (CVPR)}}. \bibinfo{pages}{1--8}.
\newblock


\bibitem[\protect\citeauthoryear{Viola and Jones}{Viola and Jones}{2004}]%
        {viola01cv}
\bibfield{author}{\bibinfo{person}{Paul Viola} {and} \bibinfo{person}{Michael
  Jones}.} \bibinfo{year}{2004}\natexlab{}.
\newblock \showarticletitle{Robust real-time object detection}.
\newblock \bibinfo{journal}{\emph{International Journal of Computer Vision}}
  \bibinfo{volume}{57}, \bibinfo{number}{2} (\bibinfo{year}{2004}),
  \bibinfo{pages}{137--154}.
\newblock


\bibitem[\protect\citeauthoryear{{Viola} and {Jones}}{{Viola} and
  {Jones}}{2001}]%
        {violajones2001}
\bibfield{author}{\bibinfo{person}{Paul {Viola}} {and}
  \bibinfo{person}{Michael~J. {Jones}}.} \bibinfo{year}{2001}\natexlab{}.
\newblock \showarticletitle{Robust real-time face detection}. In
  \bibinfo{booktitle}{\emph{Proceedings of the IEEE International Conference on
  Computer Vision (ICCV)}}, Vol.~\bibinfo{volume}{2}.
  \bibinfo{pages}{747--747}.
\newblock


\bibitem[\protect\citeauthoryear{{Vo}, {Jiang}, and {Zell}}{{Vo}
  et~al\mbox{.}}{2014}]%
        {vo2014hri}
\bibfield{author}{\bibinfo{person}{Duc~My {Vo}}, \bibinfo{person}{Lixing
  {Jiang}}, {and} \bibinfo{person}{Andreas {Zell}}.}
  \bibinfo{year}{2014}\natexlab{}.
\newblock \showarticletitle{Real time person detection and tracking by mobile
  robots using {RGB-D} images}. In \bibinfo{booktitle}{\emph{Proceedings of the
  IEEE International Conference on Robotics and Biomimetics (ROBIO)}}.
  \bibinfo{pages}{689--694}.
\newblock


\bibitem[\protect\citeauthoryear{Weiss, Bernhaupt, Tscheligi, Wollherr,
  Kuhnlenz, and Buss}{Weiss et~al\mbox{.}}{2008}]%
        {UTAUT}
\bibfield{author}{\bibinfo{person}{Astrid Weiss}, \bibinfo{person}{Regina
  Bernhaupt}, \bibinfo{person}{Manfred Tscheligi}, \bibinfo{person}{Dirk
  Wollherr}, \bibinfo{person}{Kolja Kuhnlenz}, {and} \bibinfo{person}{Martin
  Buss}.} \bibinfo{year}{2008}\natexlab{}.
\newblock \showarticletitle{A methodological variation for acceptance
  evaluation of human-robot interaction in public places}. In
  \bibinfo{booktitle}{\emph{Proceedings of the IEEE International Symposium on
  Robot and Human Interactive Communication (RO-MAN)}}.
  \bibinfo{pages}{713--718}.
\newblock


\bibitem[\protect\citeauthoryear{Yamane, Kim, and Alspach}{Yamane
  et~al\mbox{.}}{2017}]%
        {Disney}
\bibfield{author}{\bibinfo{person}{Katsu Yamane}, \bibinfo{person}{Joohyung
  Kim}, {and} \bibinfo{person}{Alexander~Nicholas Alspach}.}
  \bibinfo{year}{2017}\natexlab{}.
\newblock \bibinfo{title}{Soft Body Robot for Physical Interaction}.
\newblock \bibinfo{howpublished}{US Patent Application 20,170,095,925}.
\newblock


\bibitem[\protect\citeauthoryear{Yamazaki, Christensen, Skov, Chang, Damholdt,
  Sumioka, Nishio, and Ishiguro}{Yamazaki et~al\mbox{.}}{2016}]%
        {Hugvie}
\bibfield{author}{\bibinfo{person}{Ryuji Yamazaki}, \bibinfo{person}{Louise
  Christensen}, \bibinfo{person}{Kate Skov}, \bibinfo{person}{Chi-Chih Chang},
  \bibinfo{person}{Malene~F. Damholdt}, \bibinfo{person}{Hidenobu Sumioka},
  \bibinfo{person}{Shuichi Nishio}, {and} \bibinfo{person}{Hiroshi Ishiguro}.}
  \bibinfo{year}{2016}\natexlab{}.
\newblock \showarticletitle{Intimacy in phone conversations: anxiety reduction
  for {D}anish seniors with {H}ugvie}.
\newblock \bibinfo{journal}{\emph{Frontiers in Psychology}}
  \bibinfo{volume}{7} (\bibinfo{year}{2016}), \bibinfo{pages}{537}.
\newblock
\showISSN{1664-1078}
\urldef\tempurl%
\url{https://doi.org/10.3389/fpsyg.2016.00537}
\showDOI{\tempurl}


\bibitem[\protect\citeauthoryear{{Yurtsever}, {Lambert}, {Carballo}, and
  {Takeda}}{{Yurtsever} et~al\mbox{.}}{2020}]%
        {yurtsever2019}
\bibfield{author}{\bibinfo{person}{Ekim {Yurtsever}}, \bibinfo{person}{Jacob
  {Lambert}}, \bibinfo{person}{Alexander {Carballo}}, {and}
  \bibinfo{person}{Kazuya {Takeda}}.} \bibinfo{year}{2020}\natexlab{}.
\newblock \showarticletitle{A survey of autonomous driving: common practices
  and emerging technologies}.
\newblock \bibinfo{journal}{\emph{IEEE Access}}  \bibinfo{volume}{8}
  (\bibinfo{year}{2020}), \bibinfo{pages}{58443--58469}.
\newblock
\urldef\tempurl%
\url{https://doi.org/10.1109/ACCESS.2020.2983149}
\showDOI{\tempurl}


\end{thebibliography}
